\pdfoutput=1
\documentclass[11pt]{article}

\usepackage[]{EMNLP2022}

\usepackage{times}
\usepackage{latexsym}

\usepackage[T1]{fontenc}
\usepackage[utf8]{inputenc}

\usepackage{microtype}

\usepackage{bm}
\usepackage{graphicx}
\usepackage{amsmath}
\usepackage{multirow}
\usepackage{cleveref}
  \crefname{Figure}{Fig.}{Fig.}
\usepackage{xspace}
\usepackage{xcolor}
\usepackage{enumitem}
\usepackage{placeins}
\usepackage{booktabs}
\usepackage{soul}

\usepackage{hyperref}

\newcommand{\ours}{{QADA}\xspace}

\usepackage{inconsolata}

\title{QA Domain Adaptation using Hidden Space Augmentation and Self-Supervised Contrastive Adaptation}

\author{Zhenrui Yue\thanks{\, Both authors contributed equally to this research.} \\
  UIUC \\
  \texttt{zhenrui3@illinois.edu} \\\And
  Huimin Zeng\footnotemark[1] \\
  UIUC \\
  \texttt{huiminz3@illinois.edu} \\\And
  Bernhard Kratzwald \\
  EthonAI \\
  \texttt{bernhard.kratzwald@ethon.ai}
  \AND
  Stefan Feuerriegel \\
  LMU Munich \\
  \texttt{feuerriegel@lmu.de} \\\And
  Dong Wang \\
  UIUC \\
  \texttt{dwang24@illinois.edu}}

\begin{document}
\maketitle
\begin{abstract}
Question answering (QA) has recently shown impressive results for answering questions from customized domains. Yet, a common challenge is to adapt QA models to an unseen target domain. In this paper, we propose a novel self-supervised framework called \ours for QA domain adaptation. \ours introduces a novel data augmentation pipeline used to augment training QA samples. Different from existing methods, we enrich the samples via hidden space augmentation. For questions, we introduce multi-hop synonyms and sample augmented token embeddings with Dirichlet distributions. For contexts, we develop an augmentation method which learns to drop context spans via a custom attentive sampling strategy. Additionally, contrastive learning is integrated in the proposed self-supervised adaptation framework \ours. Unlike existing approaches, we generate pseudo labels and propose to train the model via a novel attention-based contrastive adaptation method. The attention weights are used to build informative features for discrepancy estimation that helps the QA model separate answers and generalize across source and target domains. To the best of our knowledge, our work is the first to leverage hidden space augmentation and attention-based contrastive adaptation for self-supervised domain adaptation in QA. Our evaluation shows that \ours achieves considerable improvements on multiple target datasets over state-of-the-art baselines in QA domain adaptation.
\end{abstract}

\section{Introduction}
\label{sec:introduction}

Question answering~(QA) is the task of finding answers for a given context and a given question. QA models are typically trained using data triplets consisting of context, question and answer. In the case of extractive QA, answers are represented as subspans in the context defined by a start position and an end position, while question and context are given as running text~\cite[e.g.,][]{seo2016bidirectional, chen-etal-2017-reading, devlin-etal-2019-bert, kratzwald-etal-2019-rankqa}.

A common challenge in extractive QA is that QA models often suffer from performance deterioration upon deployment and thus make mistakes for user-generated inputs. The underlying reason for such deterioration can be traced back to the domain shift between training data (from the source domain) and test data (from the target domain)~\cite{fisch-etal-2019-mrqa, miller2020effect, zeng2022attacking}. 

Existing approaches to address domain shifts in extractive QA can be grouped as follows. One approach is to include labeled target examples or user feedback during training~\cite{daume-iii-2007-frustratingly, kratzwald2019learning, kratzwald-etal-2020-learning, kamath-etal-2020-selective}. Another approach is to generate labeled QA samples in the target domain
% , which are the used additionally 
for training~\cite{lee-etal-2020-generating, yue2021cliniqg4qa, yue-etal-2022-synthetic}. However, these approaches typically require large amounts of annotated data or extensive computational resources. As such, they tend to be ineffective in adapting existing QA models to an unseen target domain~\cite{fisch-etal-2019-mrqa}. Only recently, a contrastive loss has been proposed to handle domain adaptation in QA~\cite{yue-etal-2021-contrastive}. 

Several approaches have been used to address issues related to insufficient data and generalization in NLP tasks, yet outside of QA.
For example, augmentation in the hidden space encourages more generalizable features for training~\cite{verma2019manifold, chen-etal-2020-mixtext, chen-etal-2021-hiddencut}. 
For domain adaptation, there are approaches that encourage the model to learn domain-invariant features via a domain critic~\cite{lee-etal-2019-domain, cao2020unsupervised}, or adopt discrepancy regularization between the source and target domains~\cite{kang2019contrastive, yue2022contrastive}. However, to the best of our knowledge, no work has attempted to build a smooth and generalized feature space via hidden space augmentation and self-supervised domain adaption.

In this paper, we propose a novel \emph{self-supervised QA domain adaptation} framework for extractive QA called \ours. Our \ours framework is designed to handle domain shifts and should thus answer out-of-domain questions. \ours has three stages, namely pseudo labeling, hidden space augmentation and self-supervised domain adaptation. First, we use pseudo labeling to generate and filter labeled target QA data. Next, the augmentation component integrates a novel pipeline for data augmentation to enrich training samples in the hidden space. For questions, we build upon multi-hop synonyms and introduce Dirichlet neighborhood sampling in the embedding space to generate augmented tokens. For contexts, we develop an attentive context cutoff method which learns to drop context spans via a sampling strategy using attention scores. Third, for training, we propose to train the QA model via a novel attention-based contrastive adaptation. Specifically, we use the attention weights to sample informative features that help the QA model separate answers and generalize across the source and target domains.

\textbf{Main contributions} of our work are:\footnote{The code for our \ours framework is publicly available at https://github.com/Yueeeeeeee/Self-Supervised-QA.}
\begin{enumerate}[leftmargin=15pt] %,itemsep=3px,nolistsep]
\item We propose a novel, self-supervised framework called \ours for domain adaptation in QA. \ours aims at answering out-of-domain question and should thus handle the domain shift upon deployment in an unseen domain.
\item To the best of our knowledge, \ours is the first work in QA domain adaptation that (i)~leverages hidden space augmentation to enrich training data; and (ii)~integrates attention-based contrastive learning for self-supervised adaptation.
\item We demonstrate the effectiveness of \ours in an unsupervised setting where target answers are not accessible. Here, \ours can considerably outperform state-of-the-art baselines on multiple datasets for QA domain adaptation.
\end{enumerate}

\section{Related Work}
\label{sec:related_work}

Extractive QA has achieved significantly progress recently~\cite{devlin-etal-2019-bert, kratzwald-etal-2019-rankqa, lan2019albert, zhang2020retrospective}. Yet, the accuracy of QA models can drop drastically under domain shifts; that is, when deployed in an unseen domain that differs from the training distribution~\cite{fisch-etal-2019-mrqa, talmor-berant-2019-multiqa}. 

To overcome the above challenge, various approaches for QA domain adaptation have been proposed, which can be categorized as follows. (1)~\mbox{(Semi-)}supervised adaptation uses partially labeled data from the target distribution for training~\cite{yang-etal-2017-semi, kratzwald.2019a, yue-etal-2022-synthetic}. (2)~Unsupervised adaptation with question generation refers to settings where only context paragraphs in the target domain are available, QA samples are generated separately to train the QA model~\cite{shakeri-etal-2020-end, yue-etal-2021-contrastive}. (3)~Unsupervised adaptation has access to context and question information from the target domain, whereas answers are unavailable~\cite{chung-etal-2018-supervised, cao2020unsupervised, yue-etal-2022-domain}. In this paper, we focus on the third category and study the problem of unsupervised QA domain adaptation.

\textbf{Domain adaptation for QA}: Several approaches have been developed to generate synthetic QA samples via question generation (QG) in an end-to-end fashion (i.e., seq2seq)~\cite{du-etal-2017-learning, sun-etal-2018-answer}. Leveraging such samples from QG can also improve the QA performance in out-of-domain distributions~\cite{golub-etal-2017-two, tang2017question, tang-etal-2018-learning, lee-etal-2020-generating, shakeri-etal-2020-end, yue-etal-2022-synthetic, zeng2022unsupervised}. Given unlabeled questions, there are two main approaches: domain adversarial training can be applied to reduce feature discrepancy between domains~\cite{lee-etal-2019-domain, cao2020unsupervised}, while contrastive adaptation minimizes the domain discrepancy using maximum mean discrepancy (MMD)~\cite{yue-etal-2021-contrastive, yue-etal-2022-domain}. We later use the idea from contrastive learning but tailor it carefully for our adaptation framework.

\textbf{Data augmentation for NLP}: Data augmentation for NLP aims at improving the language understanding with \emph{diverse} data samples. One approach is to apply token-level augmentation and enrich the training data with simple techniques (e.g., synonym replacement, token swapping, etc.)~\cite{wei-zou-2019-eda} or custom heuristics~\cite{mccoy-etal-2019-right}. Alternatively, augmentation can be done in the hidden space of the underlying model~\cite{chen-etal-2020-mixtext}. For example, one can drop partial spans in the hidden space, which aids generalization performance under distributional shifts~\cite{chen-etal-2021-hiddencut} but in NLP tasks outside of QA. To the best of our knowledge, we are the first to propose a hidden space augmentation pipeline tailored for QA data in which different strategies are combined for question and context augmentation.

\begin{figure*}[t]
    \centering
    \includegraphics[trim=0 6.5cm 3.5cm 0.5cm, clip, width=1.0\textwidth]{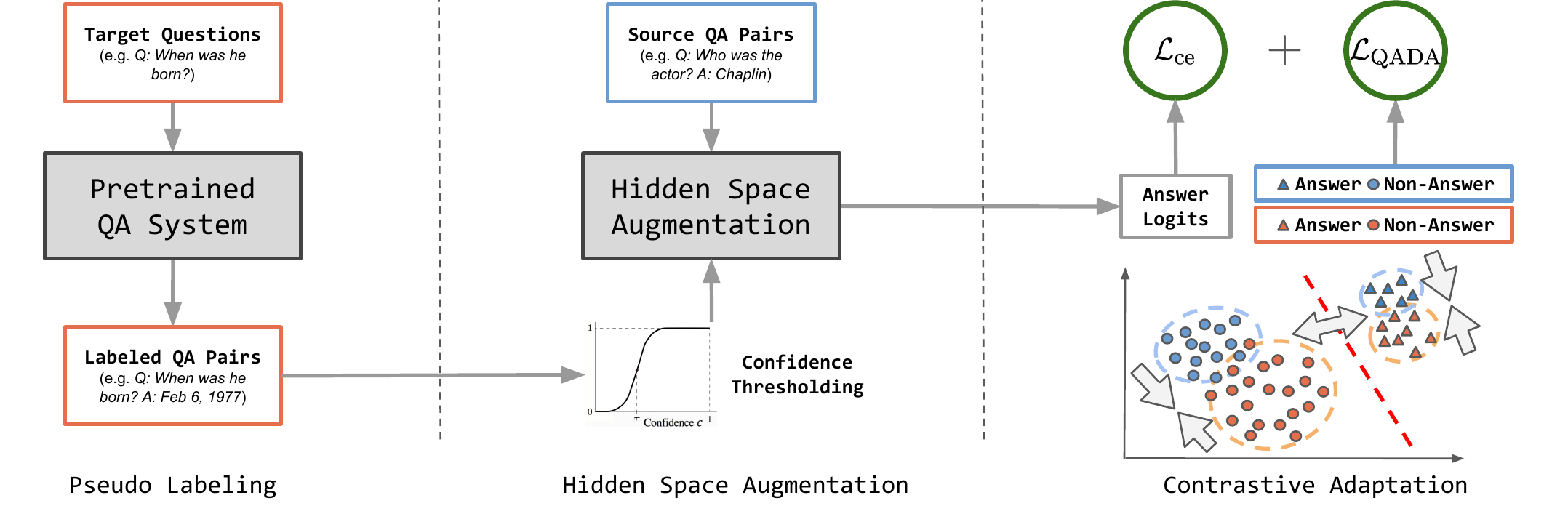}
    \caption{Overview of our proposed \ours framework. \ours generates pseudo labels for the unlabeled target data, followed by hidden space augmentation. The QA model is trained via attention-based contrastive adaptation.}
    \vspace{-8pt}
    \label{fig:framework}
\end{figure*}

\textbf{Contrastive learning for domain adaptation}: Contrastive learning is used to minimize distances of same-class samples and maximize discrepancy among classes~\cite{hadsell2006dimensionality}. For this, different metrics are adopted to measure pair-wise distances (e.g., triplet loss) or domain distances with MMD~\cite{cheng2016person, schroff2015facenet}. Contrastive learning can also be used for domain adaptation by reducing the domain discrepancy: this ``pulls together'' intra-class features and ``pushes apart'' inter-class representations. Here, several applications are in computer vision~\citep{kang2019contrastive}. In QA domain adaptation, contrastive learning was applied with averaged token features to separate answer tokens and minimize the discrepancy between source and target domain~\cite{yue-etal-2021-contrastive, yue-etal-2022-domain}. However, our work is different in that we introduce a novel \emph{attention-based} strategy to construct more informative features for discrepancy estimation and contrastive adaptation.  
\section{Setup}
\label{sec:preliminaries}

We consider the following problem setup for QA domain adaptation, where labeled source data and \emph{unlabeled} target data are available for training. Our goal is to train a QA model $\bm{f}$ that maximizes the performance in the target domain using both source data and unlabeled target data~\cite{cao2020unsupervised, shakeri-etal-2020-end, yue-etal-2021-contrastive, yue-etal-2022-domain}.  

\textbf{Data}: Our research focuses on question answering under domain shift. Let $\bm{\mathcal{D}}_{s}$ denote the source domain, and let $\bm{\mathcal{D}}_{t}$ denote the (different) target domain. Then, labeled data from the source domain can be used for training, while, upon deployment, it should perform well on the data from the target domain. Specifically, training is two-fold: we first pretrain a QA model on the source domain $\bm{\mathcal{D}}_{s}$ and, following this, the pretrained QA model is adapted to the target domain $\bm{\mathcal{D}}_{t}$. The input data to each domain is as follows:
\begin{itemize}[leftmargin=10pt]
    \item \emph{Labeled source data}: Training data is provided by labeled QA data $\bm{X}_{s}$ from the source domain $\bm{\mathcal{D}}_{s}$. Here, each sample $(\bm{x}_{s,c}^{(i)}, \bm{x}_{s,q}^{(i)}, \bm{x}_{s,a}^{(i)}) \in \bm{X}_{s}$ is a triplet comprising a context $\bm{x}_{s,c}^{(i)}$, a question $\bm{x}_{s,q}^{(i)}$, and an answer $\bm{x}_{s,a}^{(i)}$. As we consider extractive QA, the answer is represented by the start and end position in the context.
    \item \emph{Unlabeled target data}: We assume partial access to data from the target domain $\bm{\mathcal{D}}_{t}$, that is, only contexts and unlabeled questions. The contexts and questions are first used for pseudo labeling, followed by self-supervised adaptation. Formally, we refer to the contexts and questions via $\bm{x}_{t,c}^{(i)}$ and $\bm{x}_{t,q}^{(i)}$, with $(\bm{x}_{t,c}^{(i)}, \bm{x}_{t,q}^{(i)}) \in \bm{X}_{t}^{'}$ where $\bm{X}_{t}^{'}$ is the unlabeled data from the target domain. 
\end{itemize}

\noindent
\textbf{Model:} The QA model can be represented with function $\bm{f}$. $\bm{f}$ takes both a question and context as input and predicts an answer, i.e., $\bm{x}_{a}^{(i)} = \bm{f}(\bm{x}_{q}^{(i)}, \bm{x}_{c}^{(i)})$. Upon deployment, our goal is to maximize the model performance on $\bm{X}_{t}$ in the target domain $\bm{\mathcal{D}}_{t}$. Mathematically, this corresponds to the optimization of $\bm{f}$ over target data $\bm{X}_{t}$:
\begin{equation}
    \min_{\substack{\bm{f}}} \mathcal{L}_{\mathrm{ce}}(\bm{f}, \bm{X}_{t}),
\end{equation}
where $\mathcal{L}_{\mathrm{ce}}$ is the cross-entropy loss.
\section{The \ours Framework}
\label{sec:method}

\subsection{Overview}

Our proposed \ours framework has three stages to be performed in each epoch (see Fig.~\ref{fig:framework}): (1)~\textbf{pseudo labeling}, where pseudo labels are generated for the unlabeled targeted data; (2)~\textbf{hidden space augmentation}, in which the proposed augmentation strategy is leveraged to generated virtual examples in the feature space; and (3)~\textbf{contrastive adaptation} that minimizes domain discrepancy to transfer source knowledge to the target domain.

To address the domain shift upon deployment, we use the aforementioned stages as follows. In the first stage, we generate pseudo labels for the unlabeled target data $\bm{X}_{t}^{'}$.
Next, we enrich the set of training data via hidden space augmentation. In the adaptation stage, we train the QA model using both the source and the target data with our attention-based contrastive adaptation. We summarize the three stages in the following:
\begin{enumerate}[leftmargin=15pt]
    \item \emph{Pseudo labeling}: First, we build labeled target data $\hat{\bm{X}}_{t}$ via pseudo labeling. Formally, a source-pretrained QA model $\bm{f}$ generates a (pseudo) answer $\bm{x}_{t,a}^{(i)}$ for context $\bm{x}_{t,c}^{(i)}$ and question $\bm{x}_{t,q}^{(i)}$, $i = 1, \ldots$ Each sample $\bm{x}_{t}^{(i)} \in \hat{\bm{X}}_{t}$ now contains the original context, the original question, and a predicted answer. We additionally apply confidence thresholding to filter the pseudo labels.
    \item \emph{Hidden space augmentation}: The QA model $\bm{f}$ takes a question and context pair as input.
    For questions, we perform Dirichlet neighborhood sampling in the word embedding layer to generate diverse, yet consistent query information. We also apply a context cutoff in the hidden space after transformer layers to reduce the learning of redundant domain information.
    \item \emph{Contrastive adaptation}: We train the QA model $\bm{f}$ with the source data $\bm{X}_{s}$ from the source domain $\bm{\mathcal{D}}_{s}$ \underline{and} the target data $\hat{\bm{X}}_{t}$ with pseudo labels from the previous stage. We impose regularization on the answer extraction and further minimize the discrepancy between the source and target domain, so that the learned features generalize well to the target domain.
\end{enumerate}

\subsection{Pseudo Labeling}

Provided with the access to labeled source data, we first pretrain the QA model to answer questions in the source domain. Then, we can use the pretrained model to predict target answers for self-supervised adaptation~\cite{cao2020unsupervised}. The generated pseudo labels are filtered via confidence thresholding, in which the target samples above confidence threshold $\tau$ ($=0.6$ in our experiments) are preserved for the later stages. We repeat the pseudo-labeling step in each epoch to dynamically adjust the target distribution used for self-supervised adaptation.

The QA model $\bm{f}$ is pretrained on the source dataset $\bm{X}_{s}$ via a cross-entropy loss $\mathcal{L}_{\mathrm{ce}}$, i.e., $\min_{\substack{\bm{f}}} \mathcal{L}_{\mathrm{ce}}(\bm{f}, \bm{X}_{s})$. When selecting QA pairs from the target domain, we further use confidence thresholding for filtering and, thereby, build a subset of target data with pseudo labels $\hat{\bm{X}}_{t}$, i.e.,
\begin{equation}
\begin{split}
    \hat{\bm{X}}_{t} &= \big\{ (\bm{x}_{t,c}^{(i)}, \bm{x}_{t,q}^{(i)}, \bm{f}(\bm{x}_{t,c}^{(i)}, \bm{x}_{t,q}^{(i)}) \;\; \big| \\
    &\sigma(\bm{f}(\bm{x}_{t,c}^{(i)}, \bm{x}_{t,q}^{(i)})) \geq \tau, (\bm{x}_{t,c}^{(i)}, \bm{x}_{t,q}^{(i)}) \in \bm{X}_{t}^{'} \big\},
\end{split}
\end{equation}
where $\sigma$ computes the output answer confidence (i.e, softmax function).

\subsection{Hidden Space Augmentation}

We design a data augmentation pipeline to enrich the training data based on the generated QA pairs. The augmentation pipeline is divided into two parts: (i)~\emph{question augmentation} via Dirichlet neighborhood sampling in the embedding layer and (ii)~\emph{context augmentation} with attentive context cutoff in transformer layers. Both are described below.

\begin{figure}[t]
    \centering
    \includegraphics[trim=0 5.5cm 15cm 0, clip, width=1.0\linewidth]{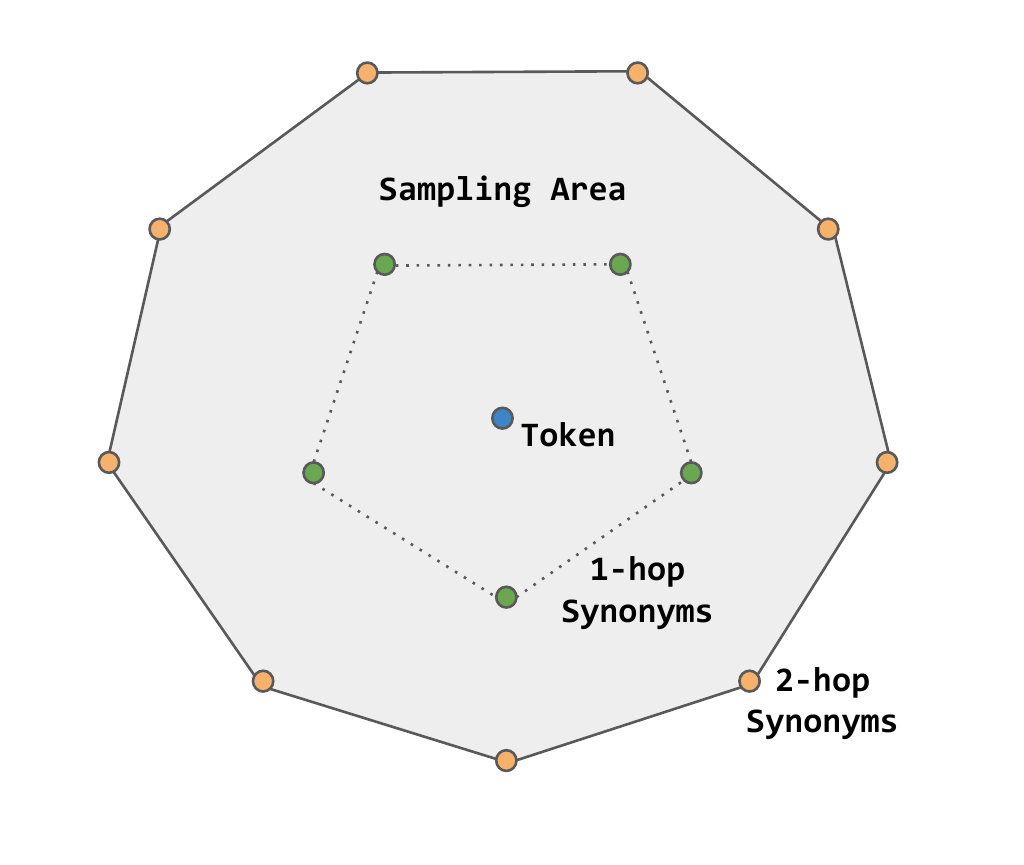}
    \caption{The proposed Dirichlet neighborhood sampling in \ours. We sample coefficients from a Dirichlet distribution and represent the augmented embedding as the linear combination of multi-hop synonyms.} 
    \vspace{-8pt}
    \label{fig:dirichlet}
\end{figure}

\textbf{Question augmentation}: To perform augmentation of questions, we propose \emph{Dirichlet neighborhood sampling} (see Fig.~\ref{fig:dirichlet}) to sample synonym replacements on certain proportion of tokens, such that the trained QA model captures different patterns of input questions. Dirichlet distributions have been previously applied to find adversarial examples~\cite{zhou-etal-2021-defense, yue2022defending}; however, different from such methods, we propose to perform question augmentation in the embedding layer. We first construct the multi-hop neighborhood for each token in the input space. Here, 1-hop synonyms can be derived from a synonym dictionary, while 2-hop synonyms can be extended from 1-hop synonyms (i.e., the synonyms of 1-hop synonyms). 

For each token, we compute a convex hull spanned by the token and its multi-hop synonyms (i.e., vertices), as shown in Fig.~\ref{fig:dirichlet}. The convex hull is used as the sampling area of the augmented token embedding, where the probability distribution in the sampling area can be computed using a Dirichlet distribution. That is, the sampled token embedding is represented as the linear combinations of vertices in the convex hull. Formally, for a token $x$ and the set of its multi-hop synonyms $\mathcal{C}_{x}$, we denote the coefficients of the linear combination by $\bm{\eta}_{x} = [ \eta_{x,1}, \eta_{x,2}, \ldots, \eta_{x,|\mathcal{C}_{x}|} ]$. We sample coefficients $\bm{\eta}_{x}$ from a Dirichlet distribution:
\begin{equation}
    \bm{\eta}_{x} \sim \mathrm{Dirichlet}(\alpha_1, \alpha_2, \ldots, \alpha_{|\mathcal{C}_{x}|}),
\end{equation}
where $\alpha$ values are selected differently for the original token and its multi-hop synonyms. Using the sampled $\bm{\eta}_{x}$, we can compute the augmented token embedding with the embedding function $\bm{f}_{e}$ via
\begin{equation}
    \bm{f}_{e}(\bm{\eta}_{x}) = \sum_{j \in \mathcal{C}_{x}}^{|\mathcal{C}_{x}|} \eta_{x,j} \, \bm{f}_{e}(j).
\end{equation}

Dirichlet distributions are multivariate probability distributions with $\sum \bm{\eta}_{x} = 1$ and $\bm{\eta}_{x, j} \geq 0, \, \forall j \in \mathcal{C}_{x}$. The augmented embedding is therefore a linear combination of vertices in the convex hull. By adjusting $\alpha$ values, we can change the probability distribution in the sampling area, that is, how far the sampled embedding can travel from the original token. For example, with increasing $\alpha$ values, we can expect the sampled points approaching the center point of the convex hull.

The above augmentation is introduced in order to provide semantically diverse yet consistent questions. At the same time, by adding noise locally, it encourages the QA model to capture robust information in questions~\cite[see][]{zhou-etal-2021-defense}. We control the question augmentation by a token augmentation ratio, $\zeta$, to determine the percentage of tokens within questions that are augmented.\footnote{We considered using the above embedding-level augmentation for contexts but eventually discarded this idea: (1)~embedding augmentation undermines the original text style and the underlying domain characteristics; and (2)~token changes for contexts are likely to cause shifts among answer spans.}

\begin{figure}[t]
    \centering
    \includegraphics[trim=0 5.5cm 15cm 0, clip, width=1.0\linewidth]{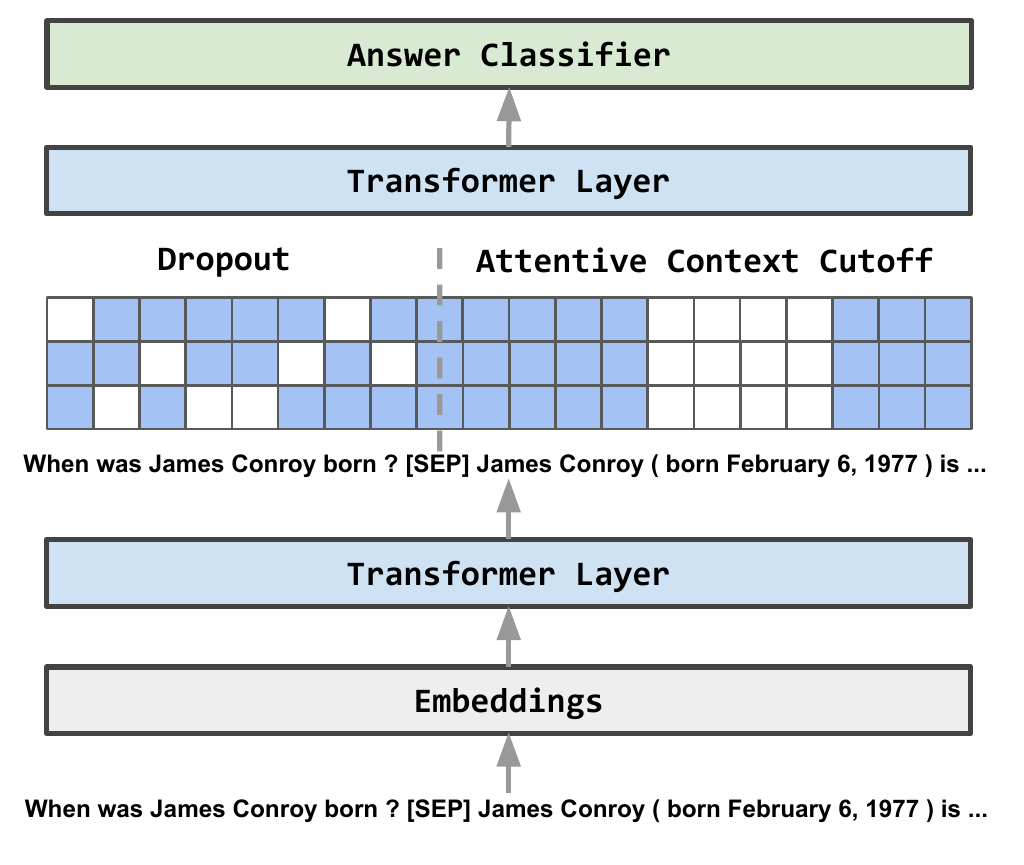}
    \caption{The proposed context cutoff in \ours. Context cutoff drops context subspans in the length dimension and is performed after transformer layers via an attentive sampling strategy.} 
    \vspace{-8pt}
    \label{fig:cutoff}
\end{figure}

\textbf{Context augmentation}: For contexts, we adopt augmentation in the hidden space of the transformer layers instead of the embedding layer. Here, we propose to use an \emph{attentive context cutoff} in the hidden space. Specifically, we zero out sampled context spans in the hidden space after each transformer layer in the QA model. This is shown in Fig.~\ref{fig:cutoff}, where all hidden states in the selected span along the input length are dropped (i.e., setting values to zero as shown by the white color). Thereby, our cutoff forces the QA model to attend to context information that is particularly relevant across all input positions and thus hinders it from learning redundant domain information. 

Formally, our attentive sampling strategy learns to select cutoff spans: we compute a probability distribution and sample a midpoint using the attention weights $A \in \bm{R}^{H \times L_{c} \times L_{c}}$ in the context span from the previous transformer layer. The probability of the $i$-th token $p_{i}$ is computed via
\begin{equation}
    p_{i} = \sigma\bigg(\frac{1}{H}\sum_{j}^{H} \bigg(\sum_{k}^{L_{c}} A_{j, k} \bigg) \bigg)_{i},
\end{equation}
where $H$ is the number of attention heads, $L_{c}$ is the context length, and $\sigma$ denotes the softmax function. 
Once the cutoff midpoint is sampled, we introduce a context cutoff ratio, $\varphi$, as a hyperparameter. It determines the cutoff length (as compared to length of the original context). We avoid context cutoff in the final transformer layer to prevent important answer features from being zeroed out.

Eventually, the above procedure of question augmentation should improve the model capacity in question understanding. Combined with context cutoff, the QA model is further forced to attend context information globally in the hidden space. This thus encourages the QA model to reduce redundancy and capture relevant information, i.e., from all context positions using self-attention. 
 
\subsection{Contrastive Adaptation}

To adapt the QA model to the target domain, we develop a tailored attention-based contrastive adaptation. Here, our idea is to regularize the intra-class discrepancy for knowledge transfer and increase the inter-class discrepancy for answer extraction. We consider answer tokens and non-answer tokens as different classes~\cite{yue-etal-2021-contrastive}.

\textbf{Loss:} We perform contrastive adaptation to reduce the intra-class discrepancy between source and target domains. We also maximize the inter-class distances between answer tokens and non-answer tokens to separate answer spans. For a mixed batch $\bm{X}$ with $\bm{X}_{s}$ and $\bm{X}_{t}$ representing the subset of source and target samples, our contrastive adaptation loss is
\begin{equation}
  \begin{aligned}
    \mathcal{L}_{\mathrm{\ours}} &= \mathcal{D}(\bm{X}_{s,a}, \bm{X}_{t,a}) + \mathcal{D}(\bm{X}_{s,n}, \bm{X}_{t,n}) \\
    &- \mathcal{D}(\bm{X}_{a}, \bm{X}_{n}) \text{\,\,\, with} \\
    \mathcal{D} &= \frac{1}{|\bm{X}_{s}||\bm{X}_{s}|} \sum_{i=1}^{|\bm{X}_{s}|} \sum_{j=1}^{|\bm{X}_{s}|} k(\phi(\bm{x}_\mathrm{s}^{(i)}), \phi(\bm{x}_\mathrm{s}^{(j)})) \\
    &+ \frac{1}{|\bm{X}_{t}||\bm{X}_{t}|} \sum_{i=1}^{|\bm{X}_{t}|} \sum_{j=1}^{|\bm{X}_{t}|} k(\phi(\bm{x}_\mathrm{t}^{(i)}), \phi(\bm{x}_\mathrm{t}^{(j)})) \\
    &- \frac{2}{|\bm{X}_{s}||\bm{X}_{t}|} \sum_{i=1}^{|\bm{X}_{s}|} \sum_{j=1}^{|\bm{X}_{t}|} k(\phi(\bm{x}_\mathrm{s}^{(i)}), \phi(\bm{x}_\mathrm{t}^{(j)})),
  \label{eq:contrastive_loss}
  \end{aligned}
\end{equation}
where $\bm{X}_{a}$ represents answer tokens, $\bm{X}_{n}$ represents non-answer tokens in $\bm{X}$. $\bm{x}_\mathrm{s}^{(i)}$ is the $i$-th sample from $\bm{X}_{s}$, and $\bm{x}_\mathrm{t}^{(j)}$ is the $j$-th sample from $\bm{X}_{t}$. $\mathcal{D}$ computes the MMD distance with empirical kernel mean embeddings and Gaussian kernel $k$ using our scheme below. In $\mathcal{L}_{\mathrm{\ours}}$, the first two terms reduce the intra-class discrepancy (\emph{discrepancy term}), while the last term maximizes the distance of answer tokens to other tokens, thereby improving answer extraction (\emph{extraction term}). 

\textbf{MMD:} The maximum mean discrepancy (MMD) computes the proximity between probabilistic distributions in the reproducing kernel Hilbert space $\mathcal{H}$ using drawn samples~\cite{gretton2012kernel}. 
% In our implementation, we compute the discrepancy between the source and target batches with empirical kernel mean embeddings. 
In previous research~\cite{yue-etal-2021-contrastive}, the MMD distance $\mathcal{D}$ was computed using the BERT encoder. However, simply using $\phi$ as in previous work would return the \emph{averaged} feature of relevant tokens in the sample rather than more \emph{informative} tokens (i.e., tokens near the decision boundary which are ``harder'' to classify). 

Unlike previous methods, we design an attention-based sampling strategy. First, we leverage the attention weights $A \in \bm{R}^{H \times L_{\bm{x}} \times L_{\bm{x}}}$ of input $\bm{x}$ using the encoder of the QA model. Based on this, we compute a probability distribution for tokens of the relevant class (e.g., non-answer tokens) using the softmax function $\sigma$ and sample an index. The corresponding token feature from the QA encoder is used as the class feature, i.e.,
\begin{equation}
  \phi(\bm{x}) = \bm{f}_{\mathrm{enc}}(\bm{x})_{i} \text{ with } i \sim \sigma\bigg(\frac{1}{H}\sum_{j}^{H} \sum_{k}^{L_{\bm{x}}} A_{j, k} \bigg) ,
  \label{eq:attn-token-features}
\end{equation}
where $\bm{f}_{\mathrm{enc}}$ is the encoder of the QA model. As a result, features are sampled proportionally to the attention weights. This should reflect more representative information of the token class for discrepancy estimation. We apply the attention-based sampling to both answer and non-answer features. 

\begin{figure}[t]
    \centering
    \includegraphics[trim=0 7cm 15.5cm 0, clip, width=1.0\linewidth]{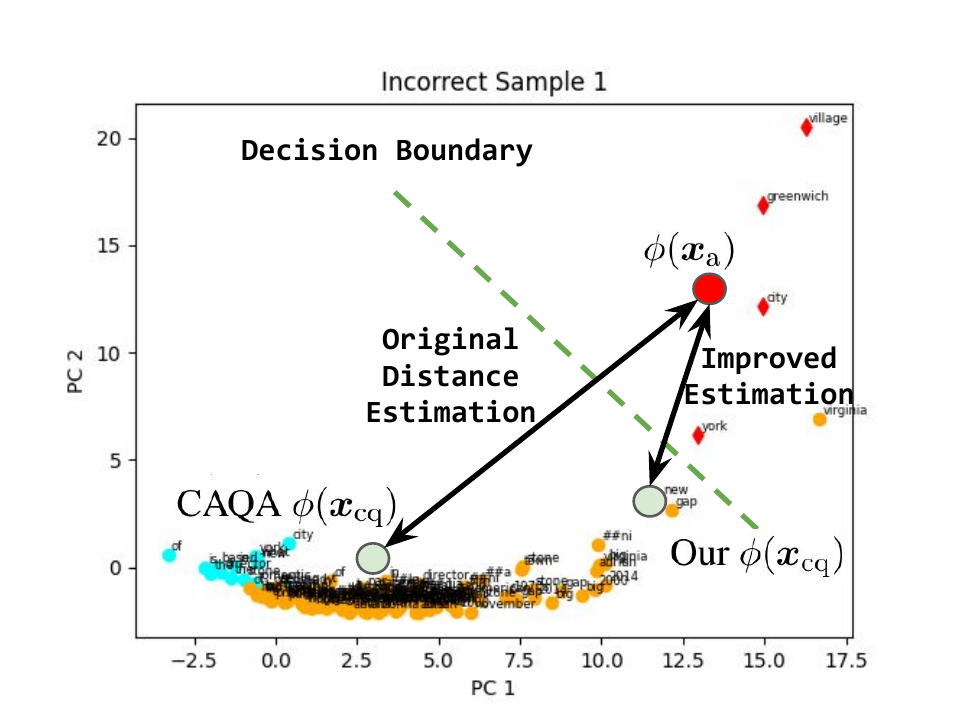}
    \caption{Illustration of QA samples in feature space. The token features are obtained from the last layer in BERT and visualized using principle component analysis (PCA). Question tokens are in cyan, context tokens in orange, and answer tokens in red. We compared the feature mappings $\phi$ between (i)~\citet{yue-etal-2021-contrastive} (called CAQA) vs. (ii)~our \ours framework.}
    \vspace{-8pt}
    \label{fig:distance-estimation}
\end{figure}

\begin{table*}[t]
\centering
\begin{tabular}{lccccc}
\toprule
\multirow{2}{*}{\textbf{Model}}                          & \textbf{HotpotQA}       & \textbf{NaturalQ.}      & \textbf{NewsQA}         & \textbf{SearchQA}       & \textbf{TriviaQA}        \\
                                                         & EM / F1                 & EM / F1                 & EM / F1                 & EM / F1                 & EM / F1                 \\ \midrule
\multicolumn{6}{c}{(I) Zero-shot target performance}                                                                                                                                        \\ \midrule
BERT                                                     & 43.34/60.42             & 39.06/53.7              & 39.17/56.14             & 16.19/25.03             & 49.70/59.09             \\ \midrule
\multicolumn{6}{c}{(II) Target performance w/ domain adaptation}                                                                                                                            \\ \midrule
DAT~\cite{lee-etal-2019-domain}                          & 44.25/61.10             & 44.94/58.91             & 38.73/54.24             & 22.31/31.64             & 49.94/59.82             \\
CASe~\cite{cao2020unsupervised}                          & 47.16/63.88             & 46.53/60.19             & 43.43/59.67             & 26.07/35.16             & 54.74/63.61             \\
CAQA~\cite{yue-etal-2021-contrastive}                    & 46.37/61.57             & \ul{48.55/62.60}        & 40.55/55.90             & \ul{36.05/42.94}        & \ul{55.17/63.23}        \\
CAQA\textsuperscript{*}~\cite{yue-etal-2021-contrastive} & \ul{48.52/64.76}        & 47.37/60.52             & \ul{44.26/60.83}        & 32.05/41.07             & 54.30/62.98             \\
\ours (ours)                                             & \textbf{50.80/65.75}    & \textbf{52.13/65.00}    & \textbf{45.64/61.84}    & \textbf{40.47/48.76}    & \textbf{56.92/65.86}    \\ \midrule
\multicolumn{6}{c}{(III) Target performance w/ supervised training}                                                                                                                         \\ \midrule
BERT w/ 10k Annotations                                  & 49.52/66.56             & 54.88/68.10             & 45.92/61.85             & 60.20/66.96             & 54.63/60.73             \\
BERT w/ All Annotations                                  & 57.96/74.76             & 67.08/79.02             & 52.14/67.46             & 71.54/77.77             & 64.51/70.27             \\ \bottomrule
\end{tabular}
\caption[Main results]{Main results of QA adaptation performance on target dataset.}
\vspace{-8pt}
\label{tab:main-results}
\end{table*}

\textbf{Illustration:} We visualize an illustrative QA sample in Fig.~\ref{fig:distance-estimation} to explain the advantage of our attention-based sampling for domain discrepancy estimation. We visualize all token features and then examine the extraction term from Eq.~\ref{eq:contrastive_loss}. We further show the feature mapping $\phi$ from~\citet{yue-etal-2021-contrastive}, which, different from ours, returns the \emph{average} feature. In contrast, our $\phi$ focuses on the estimation of more informative distances. 
As a result, our proposed attention-based sampling strategy is more likely to sample ``harder'' context tokens. These are closer to the decision boundary, as such token positions have higher weights in $A$. Owing to our choice of $\phi$, \ours improves the measure of answer-context discrepancy and, therefore, is more effective in separating answer tokens.

\subsection{Learning Algorithm}

We incorporate the contrastive adaptation loss from Eq.~\ref{eq:contrastive_loss} into the original training objective. This gives our overall loss
\begin{equation}
  \mathcal{L} = \mathcal{L}_{\mathrm{ce}} + \lambda \mathcal{L}_{\mathrm{\ours}},
  \label{eq:overall-objective}
\end{equation}
where $\lambda$ is a weighting factor for the contrastive loss.
\section{Experiments}
\label{sec:experiments}

\textbf{Datasets}: We use the following datasets (see \Cref{appendix:datasets} for details):
\begin{itemize}[leftmargin=10pt]
\item For the \emph{source domain} $\bm{\mathcal{D}}_{s}$, we use SQuAD v1.1~\cite{rajpurkar-etal-2016-squad}. 
\item For \emph{target domain} $\bm{\mathcal{D}}_{t}$, we select MRQA Split I~\cite{fisch-etal-2019-mrqa}: HotpotQA~\cite{yang-etal-2018-hotpotqa}, Natural Questions~\cite{kwiatkowski-etal-2019-natural}, NewsQA~\cite{trischler2016newsqa}, SearchQA~\cite{dunn2017searchqa}, and \mbox{TriviaQA} \cite{joshi-etal-2017-triviaqa}. This selection makes our results comparable with other works in QA domain adaptation~\cite[e.g.,][]{lee-etal-2020-generating, shakeri-etal-2020-end, cao2020unsupervised, yue-etal-2021-contrastive, yue-etal-2022-domain}.
\end{itemize}

\textbf{Baselines:} 
% We choose baselines that are designed for QA domain adaptation (i.e., they perform the same task as our \ours). 
As a na{\"i}ve baseline, we pretrain a BERT on the source dataset as our base model and evaluate on each target dataset with zero knowledge of the target domain. In addition, we adopt three state-of-the-art baselines: domain-adversarial training (DAT)~\cite{lee-etal-2019-domain}, conditional adversarial self-training (CASe)~\cite{cao2020unsupervised}, and contrastive adaptation for QA (CAQA)~\cite{yue-etal-2021-contrastive}. For a fair comparison, we adopt both the original CAQA and CAQA with our self-supervised adaptation framework ($=$ CAQA\textsuperscript{*}).\footnote{For CAQA\textsuperscript{*}, we exclude question generation and adopt the same process of pseudo labeling and self-supervised adaptation as \ours. Different from \ours, hidden space augmentation is not applied and we use the same objective function as in the original CAQA paper.} Baseline details are reported in \Cref{appendix:baselines}.

\textbf{Training and Evaluation:} We use the proposed method to adapt the pretrained QA model, augmentation hyperparameters are tuned empirically by searching for the best combinations. To evaluate the predictions, we follow~\cite{lee-etal-2020-generating, shakeri-etal-2020-end, yue-etal-2021-contrastive} and assess the exact matches (EM) and the F1 score on the dev sets. Implementation details are in \Cref{appendix:implementation}. 
\begin{table*}[t]
\centering
\begin{tabular}{@{}lccccc@{}}
\toprule
\multirow{2}{*}{\textbf{Model}}           & \textbf{HotpotQA}    & \textbf{NaturalQ.}   & \textbf{NewsQA}      & \textbf{SearchQA}    & \textbf{TriviaQA}    \\ 
                                          & EM / F1              & EM / F1              & EM / F1              & EM / F1              & EM / F1              \\ \midrule
\ours (ours)                              & \textbf{50.80/65.75} & \textbf{52.13/65.00} & \textbf{45.64/61.84} & \textbf{40.47/48.76} & \textbf{56.92/65.86} \\
w/o Dirichlet sampling                    & 49.57/64.71          & 51.15/64.24          & 45.27/61.44          & 35.90/44.28          & 56.83/65.51          \\
w/o context cutoff                        & 50.36/65.71          & 50.30/62.98          & 45.39/61.47          & 33.94/42.43          & 56.04/64.87          \\
w/o contrastive adaptation                & 48.21/64.54          & 48.35/61.76          & 44.35/60.66          & 30.85/39.42          & 55.42/64.38          \\ \bottomrule
\end{tabular}
\caption{Ablation study on different components of \ours.}
\vspace{-8pt}
\label{tab:ablation}
\end{table*}

\section{Experimental Results}

\subsection{Adaptation Performance}

Our main results for domain adaptation are in \Cref{tab:main-results}. We distinguish three major groups: (1)~\emph{Zero-shot target performance.} Here, we report a na{\"i}ve baseline (BERT) for which the QA model is solely trained on SQuAD.
(2)~\emph{Target performance w/ domain adaptation}. This refers to the methods where domain adaptation techniques are applied. This group also includes our proposed \ours. (3)~\emph{Target performance w/ supervised training}. Here, training is done with the original target data. Hence, this reflects an ``upper bound''. 

Overall, the domain adaptation baselines are outperformed by \ours across all target datasets. Hence, this confirms the effectiveness of the proposed framework using both data augmentation and attention-based contrastive adaptation. In addition, we observe the following: (1)~All adaptation methods achieve considerable improvements in answering target domain questions compared to the na{\"i}ve baseline. (2)~\ours performs the best overall. Compared to the best baseline, \ours achieves performance improvements by $6.1\%$ and $4.9\%$ in EM and F1, respectively. (3)~The improvements with \ours are comparatively larger on HotpotQA, Natural Questions, and SearchQA ($\sim8.1\%$ in EM) in contrast to NewsQA and TriviaQA ($\sim3.1\%$ in EM). A potential reason for the gap is the limited performance of BERT in cross-sentence reasoning, where the QA model often fails to answer compositional questions in long input contexts. (4)~\ours can perform similarly or outperform the supervised training results using 10k target data. For example, \ours achieve 56.92 (EM) and 65.86 (F1) on TriviaQA in contrast to 54.63 and 60.73 of the 10k supervised results, suggesting the effectiveness of \ours.

\subsection{Ablation Study for \ours}
\label{sec:ablation_study}

We evaluate the effectiveness of the proposed \ours by performing an ablation study. By comparing the performance of \ours and CAQA\textsuperscript{*} in \Cref{tab:main-results}, we yield an ablation quantifying the gains that should be attributed to the combination of all proposed components in \ours. We find distinctive performance improvements due to our hidden space augmentation and contrastive adaptation. For example, we observe that EM performance can drop up to $20.8\%$ without \ours, suggesting clear superiority of the proposed \ours.

We further evaluate the effectiveness of the individual components in \ours. We remove the proposed Dirichlet neighborhood sampling, attentive context cutoff and attention-based contrastive adaptation in \ours separately and observe the performance changes. The results on target datasets are reported in \Cref{tab:ablation}. For all components, we observe consistent performance drops when removed from \ours. For example, the performance of \ours reduces, on average, by $3.3\%$, $4.5\%$, and $8.3\%$ in EM when we remove Dirichlet sampling, context cutoff, and contrastive adaptation, respectively. The results suggest that the combination of question and context augmentation in the hidden space is highly effective for improving QA domain adaptation. Moreover, the performance improves clearly when including our attention-based contrastive adaptation.

\begin{figure}[h]
    \centering
    \includegraphics[trim=1.5cm 0 2cm 0, clip, width=1.0\linewidth]{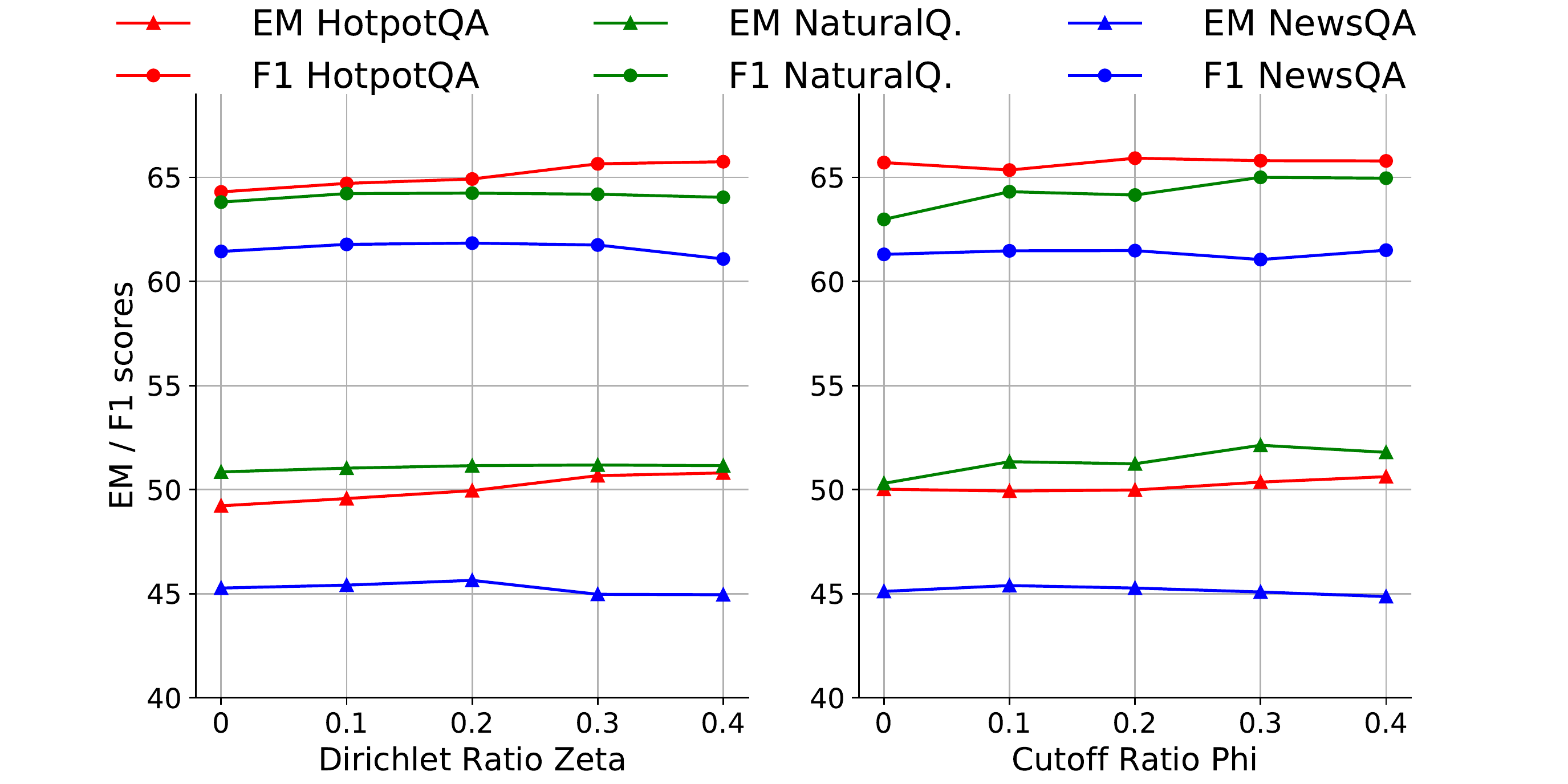}
    \caption{Sensitivity analysis for different Dirichlet sampling ratios (left) and context cutoff ratios (right).}
    \vspace{-8pt}
    \label{fig:augmentation_ratio}
\end{figure}

\subsection{Sensitivity Analysis for Hidden Space Augmentation}
\label{sec:augmentation_analysis}

Our \ours uses question augmentation (i.e., Dirichlet neighborhood sampling) and context augmentation (i.e., context cutoff), where the augmentation ratios determine the percentage of tokens that are augmented. \Cref{fig:augmentation_ratio} compares different augmentation ratios from $0$ to $0.4$ on HotpotQA, Natural Questions, and NewsQA. Overall, we observe some variations but, importantly, the performance of \ours improves adaptation performance and remains fairly robust for different nonzero ratios. Moreover, we find comparatively large improvements for HotpotQA by introducing Dirichlet neighborhood sampling ($2.5\%$ in EM), while Natural Questions benefits more from context cutoff ($3.6\%$ in EM). A potential reason for such improvements is that HotpotQA has more complex questions that need potential matching and reasoning, while Natural Questions provides longer unstructured text as contexts, thereby requiring improved understanding of long paragraphs.

\section{Conclusion}

In this paper, we propose a novel self-supervised framework called \ours for QA domain adaptation. \ours introduces: (1)~hidden space augmentation tailored for QA data to enrich target training corpora; and (2)~an attention-based contrastive adaptation to learn domain-invariant features that generalize across source and target domain. Our experiments demonstrate the effectiveness of \ours: it achieves a superior performance over state-of-the-art baselines in QA domain adaptation.

\section{Limitations}

Despite having introduced hidden space augmentation in \ours, we have not discussed different choices of $\alpha$ values for multi-hop synonyms to exploit the potential benefits of the Dirichlet distribution. For context cutoff, dropping multiple context spans in each QA example may bring additional benefits to improve context understanding and the answer extraction process of the QA model. Combined with additional question value estimation in pseudo labeling, we plan to explore such directions in adaptive QA systems as our future work.

\section*{Acknowledgments}

This research is supported in part by the National Science Foundation under Grant No. IIS-2202481, CHE-2105005, IIS-2008228, CNS-1845639, CNS-1831669. The views and conclusions contained in this document are those of the authors and should not be interpreted as representing the official policies, either expressed or implied, of the U.S. Government. The U.S. Government is authorized to reproduce and distribute reprints for Government purposes notwithstanding any copyright notation here on.

\bibliography{anthology,custom}
\bibliographystyle{acl_natbib}

\appendix
\clearpage
\section*{Appendix}

\section{Dataset Details}
\label{appendix:datasets}

For the source domain, we adopt \textbf{SQuAD v1.1}~\cite{rajpurkar-etal-2016-squad} following~\cite{cao2020unsupervised, lee-etal-2020-generating, shakeri-etal-2020-end, yue-etal-2021-contrastive}. SQuAD v1.1 is a question-answering dataset where context paragraphs originate from Wikipedia articles. The QA pairs were then annotated by crowdworkers.
  
In our experiments, we adopt all datasets from MRQA Split I~\cite{fisch-etal-2019-mrqa} for the target domains:
\begin{enumerate}[leftmargin=15pt,itemsep=3px,nolistsep]
  \item \textbf{HotpotQA} is a question-answering dataset with multi-hop questions and supporting facts to promote reasoning in QA~\cite{yang-etal-2018-hotpotqa}.
  \item \textbf{NaturalQuestions}~\cite{kwiatkowski-etal-2019-natural} builds upon real-world user questions. These were then combined with Wikipedia articles as context. The Wikipedia articles may or may not contain the answer to each question.
  \item \textbf{NewsQA}~\cite{trischler2016newsqa} provides news as contexts and challenging questions beyond simple matching and entailment.
  \item \textbf{SearchQA}~\cite{dunn2017searchqa} was built based on an existing dataset of QA pairs. The QA pairs were then extended by contexts, which were crawled through Google search.
  \item \textbf{TriviaQA}~\cite{joshi-etal-2017-triviaqa} is a question-answering dataset containing evidence information for reasoning in QA.
\end{enumerate}

\section{Baseline Details}
\label{appendix:baselines}

As a na{\"i}ve baseline, we adopt BERT (uncased base version with additional batch normalization layer) and train on the source dataset~\cite{devlin-etal-2019-bert, cao2020unsupervised}. 
Additionally, we implemented the following three baselines for unsupervised QA domain adaptation:
\begin{enumerate}
\item \textbf{Domain adversarial training (DAT)}~\cite{tzeng2017adversarial, lee-etal-2019-domain} consists of a QA system and a discriminator using \texttt{[CLS]} output in BERT. The QA system is first trained on labeled source data. Then, input data from both domains is used for domain adversarial training to learn generalized features.
\item \textbf{Conditional adversarial self-training (CASe)}~\cite{cao2020unsupervised} leverages self-training with conditional
adversarial learning across domains. CASe iteratively perform pseudo labeling and domain adversarial training to reduce domain discrepancy. We adopt the entropy weighted CASe+E in our work as baseline. 
\item \textbf{CAQA}~\cite{yue-etal-2021-contrastive} leverages QAGen-T5 for question generation but extends the learning algorithm with a contrastive loss on token-level features for generalized QA features. Specifically, CAQA uses contrastive adaptation to reduce domain discrepancy and promote answer extraction.
\item \textbf{Self-supervised contrastive adaptation for QA (CAQA\textsuperscript{*})}~\cite{yue-etal-2021-contrastive} is a modified self-supervised baseline based on CAQA. We exclude question generation and adopt the same process of pseudo labeling and self-supervised adaptation as in \ours. Unlike \ours, hidden space augmentation and attention-based contrastive loss are removed.
\end{enumerate}

\section{Implementation Details}
\label{appendix:implementation}

\textbf{QA model}: We adopt BERT with an additional batch norm layer after the encoder for QA domain adaptation, as in~\cite{cao2020unsupervised}. We first pretrain BERT with a learning rate of $3 \cdot 10^{-5}$ for two epochs and a batch size of $12$ on the source dataset. We use the AdamW optimizer with $10\%$ linear warmup. We additionally use Apex for mixed precision training.

\textbf{Adaptation}: For the baselines, we use the original BERT architecture and follow the default settings provided in the original papers. For \ours, adaptation is performed 4 epochs with the AdamW optimizer, learning rate of $2 \cdot 10^{-5}$, and 10\% proportion as warmup in each epoch (as training data changes after pseudo labeling). In the pseudo labeling stage, we perform inference on unlabeled target data and preserve the target samples with confidence above the threshold $\tau = 0.6$. For batching in self-supervised adaptation, we perform hidden space augmentation and sample 12 target examples and another 12 source examples.

\textbf{\ours}: For our experiments, the scaling factor $\lambda$ for the adaptation loss is chosen from $[0.0001, 0.0005]$ depending on the target dataset. For Dirichlet neighborhood sampling, we use $\alpha = 1$ for the original token and a decay factor of $0.1$ for multi-hop synonyms (i.e., $0.1$ for 1-hop synonyms and $0.01$ for 2-hop synonyms). For hyperparameters in hidden space augmentation, we search for a combination of question augmentation ratio $\zeta$ and context cutoff ratio $\varphi$. Specifically, we empirically search for the best combination in the range of $[0.1, 0.2, 0.3, 0.4]$ for both $\zeta$ and $\varphi$. Eventually, the best hyperparameter combination is selected. 

\end{document}